\documentclass[journal]{IEEEtran}
\usepackage{multirow}

\ifCLASSINFOpdf
   \usepackage[pdftex]{graphicx}
\else
\fi
\usepackage{amsmath}
\begin{document}
%
\title{A Survey on Natural Language Video Localization}
%
%
%

\author{Xinfang Liu,
Xiushan Nie,~\IEEEmembership{Member,~IEEE,}
        Zhifang Tan, Jie Guo,
       Yilong Yin
\thanks{Xinfang Liu and Yilong Yin are with the School of software, Shandong University, China.

Xiushan Nie, Zhifang Tan and Jie Guo are with School of Computer Science and Technology, Shandong Jianzhu University.}
}
%
%

\markboth{}%
{}
%



\maketitle

\begin{abstract}
Natural language video localization (NLVL), which aims to locate a target moment from a video that semantically corresponds to a text query, is a novel and challenging task. Toward this end, in this paper, we present a comprehensive survey of the NLVL algorithms, where we first propose the pipeline of NLVL, and then categorize them into supervised and weakly-supervised methods, following by the analysis of the strengths and weaknesses of each kind of methods. Subsequently, we present the dataset, evaluation protocols and the general performance analysis. Finally, the possible perspectives are obtained by summarizing the existing methods.
\end{abstract}


%
\IEEEpeerreviewmaketitle

\section{Introduction}
\begin{figure} 
\centering 
\includegraphics[width=0.5\textwidth]{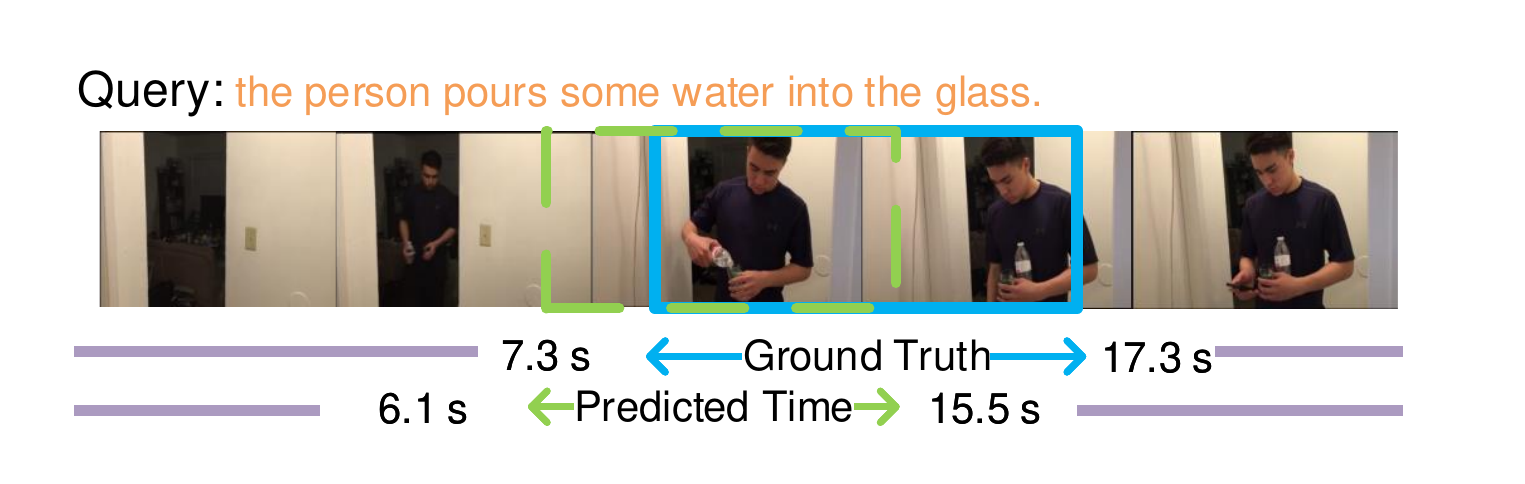}
\caption{Natural Language Video Localization aims at locating the segment that is most relevant to the query sentence.} 
\label{fig:task} 
\end{figure}

Given a video and a query sentence described in natural language form, natural language video localization (NLVL) aims at finding the segment from the video that is relevant to the query description, i.e., giving the start and end timestamps of the segment. 
Figure \ref{fig:task} shows a NLVL scenario in which one might be interested only when ``the person pours some water into the glass" in the video, where NLVL gives the moment start and end times of 6.1 s and 15.5 s, while the ground truth start and end times are 7.3 s and 17.3 s. 
NLVL is also called moment retrieval, moment localization, or video grounding.
As an emerging task,  NLVL is easily confused with
other traditional research domains. We briefly summarize the tasks of the following research domains to show how they differ from NLVL.

    \textbf{Temporal Action Localization}. It is dedicated to determining the dutaion and category in which the action occurs from the video \cite{shou2016temporal,gao2017cascaded,chao2018rethinking,buch2019end,long2019gaussian}. Specifically, what happened in this video.
    It is possible for these events to occur multiple times in the video, and for the duration of the events to be both long and short.
    Considering an action can be described precisely using the natural language, queries in NLVL usually correspond only to a certain segment of the video. 
    As a result, actions in NLVL are open-ended and not restricted to a limited number of categories as that in Temporal Action Localization.
    Since temporal action localization and NLVL have many similarities, many models or algorithms in Temporal Action Localization can be migrated to NLVL.
    
   \textbf{Video Captioning}.
   It focus on translating a video into the text based on its semantics \cite{8627985,Wang_2018_CVPR,Pan_2017_CVPR,7984828,Wang_2018_CVPR2,Shen_2017_CVPR}.
    One possible application scenario is to quickly generate a summary of the video to help the viewer determine the value of the video to save time.
    Although its mission objectives seem to be the opposite of NLVL, they share many similarities in techniques. For example, they need to process data from both video and text modals and align them semantically. Inspired by this, some
    models are trained on the basis of this reciprocal relationship, and some NLVL work borrows from the network structure in Video Captioning. As a result, some models are able to address both tasks.
    
     \textbf{Temporal Action Proposals}. It divides the video into several segments based on the semantics of different actions 
    \cite{buch2017sst,Gao_2017_ICCV,escorcia2016daps,gao2020play}. 
    It started as a part of  Temporal Action Localization, and gradually develops  into a separate branch. 
    Its original intention was to facilitate the processing of long videos. 
    By dividing the video semantically into multiple independent segments, researchers can focus on downstream tasks, such as action recognition. 
    A considerable amount of NLVL work has borrowed or directly used some Temporal Action Proposals methods.
    
It had been an almost unthinkable task to find the corresponding segment from a video
based on natural language descriptions. Later, with the development of deep learning techniques for image recognition, target detection, text processing, and many other aspects, people gradually saw hope in the field of video understanding. Recently, with the advent of computer hardware technology and visual and textual groundwork, NLVL has become possible.

If the NLVL problem can be solved perfectly, great benefits will be brought to our life. For example, it could automatically cut a compilation from many videos based on descriptions, locate the movie episode you want to watch most, and extract the most helpful segment from hours of surveillance video. In addition to its application value, research about it will also contribute to  other areas of video understanding. Therefore, it has received increasing attention from researchers around the world.

However, NLVL faces many challenges because  it  requires simultaneous processing
from both visual and textual modalities. In particular, it is intractable to align semantic information from both
visual and textual modalities. Besides, how to obtain the
final timestamp from the alignment relationship is also a tough task. Although existing approaches propose many impressive strategies to address these challenges, whether these approaches can be successfully applied to real-life situations is another matter that deserves careful consideration.  In what follows, we summarize the following evaluation aspects
to analyze the strengths and weaknesses of the existing approaches.
\begin{itemize}
    \item \textbf{Precision}. Although there has been a great deal of research to improve accuracy step by step, there is still a big gap between the current accuracy level and the actual industrial fielding needs. The core content of improving accuracy is to fully exploit the fine-grained matching relationship between the two modalities of video and natural language.
    \item \textbf{Efficiency}. Understanding the semantic information of video and text requires a large amount of computation under current conditions. However, extensive computation may  limit the scenarios in which NLVL can be used, Thus, it is also a challenge to keep the accuracy and computation cost low.
    \item \textbf{Robustness}. Is the model suitable for open scenarios?  For example, can a model pre-trained with static features accurately localize actions?  Can the model designed for current short video length datasets be easily extended to longer videos?
\end{itemize}

This survey targets at helping beginners who are interested in NLVL. It provides an overview of what has been done and what the future holds. Section II explains the basic workflow of NLVL and serves as a starting point to help the reader understand the subsequent sections. Section III introduces the usage of supervisory and weak supervisory methods in NLVL. Section IV lists the common benchmark datasets and evaluation metrics for NLVL tasks. Section V shows the performance comparison for exsiting methods. Section VI summarizes existing work and gives possible prospects for development.

\section{Pipeline}
\begin{figure*} 
\centering 
\includegraphics[width= 0.8\textwidth]{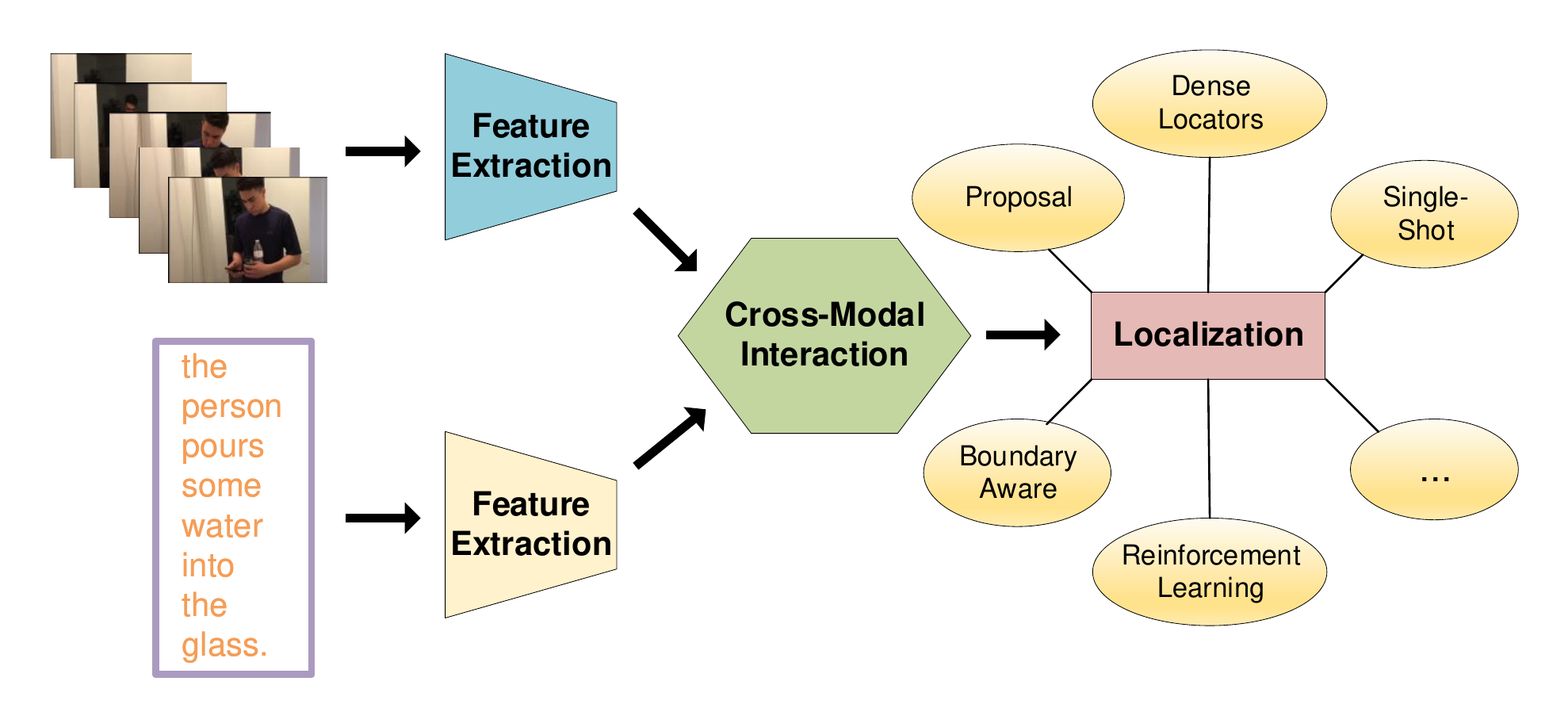}

\caption{A typical event localization network consists of three parts: 1) feature extraction; 2) feature interaction; and 3) location prediction.} 
\label{fig:pipeline} 
\end{figure*}
Figure \ref{fig:pipeline} shows a highly abstract pipeline for NLVL. Specifically, first, the video is extracted as a series of feature vectors that represent the semantic information of the video content at each timestep, while the query sentence is extracted as one feature vector that represents the information of the whole sentence or as a collection of many word feature vectors, depending on the model. Then, feature vectors from two different modalities will interact cross-modally to explore the semantic matching relations between them. Finally, localization is done based on the results of the interaction. These three steps are described in detail below.
\subsection{Feature Extraction}
\subsubsection{Video Feature Extraction}
Given the video is a superposition of multiple images, training directly from the raw videos will greatly increase the computational cost. In addition, the sizes of the existing datasets are far too small for an open NLVL task to perform robust end-to-end training. Therefore, the vast majority of the work is based on pre-extracted video features to design and test their own models rather than training from the original raw videos. 
The video feature extractors used in existing work can be broadly classified into two categories. One is based on the clip level, such as C3D \cite{tran2015learning}, I3D \cite{carreira2017quo}, etc., where the input multiple frames output a feature vector for capturing dynamic information in the video. The other is based on the frame level, such as VGG \cite{simonyan2014very}, Faster-RCNN \cite{girshick2015fast}, etc., where the feature vector outputs from one frame and it is more concerned with object information. Without causing ambiguity in the following, we consider the clip as the smallest unit of video features, and the frame is a special case of the clip.
\subsubsection{Text Feature Extraction}
Considering a query sentence is usually composed of multiple words, there are also two levels of feature representation for query. To be more specific, the first one is to extract the whole sentence feature directly \cite{kiros2015skip}, and the second one is to extract the features of words separately and then conduct further processing. However, the latter is mostly used because it is convenient to mine the information of certain keywords in the query sentence, which is usually processed by a combination of NLTK\cite{bird2009natural}+GLove\cite{pennington2014glove}+LSTM\cite{hochreiter1997long}/GRU\cite{chung2014empirical}.

\subsection{Cross-Modal Interaction}
In this paper, cross-modal interaction is  in a broad sense rather than specifically used for CMIN \cite{zhang2019cross}. The role of cross-modal interactions is to allow video features and text features to perceive each other and thus enhance useful information. This part usually harbors unique designs by researchers and is a key part of mining fine-grained matching relationships. The attention mechanism and graph neural networks are the main means to perform cross-modal interactions. However, there are also various other designs to fully communicate information between the two modalities. After the cross-modal interaction between the features of two modalities, a sequence of mixed-modal features are obtained, on which the localization module makes predictions.
Of course, some methods do not perform cross-modal interaction, instead, they perform a similarity calculation after linear or nonlinear transformation of the features, and then select the best matching segment based on the similarity.
\subsection{Localization Policies}
The localization policies is used to generate prediction bounds from the  mixed-modal features   or similarities. If the quality of cross-modal interactions determines the accuracy, then differences in localization strategies will lead to differences in efficiency and robustness.
In this study, we summarize the existing methods into the following five major categories, noting that the categories are not strictly differentiated, e.g., in reinforcement learning, candidate window ranking or regression may be performed. 
\begin{itemize}
    \item  \emph{Proposal-Based Methods.} The original video is first divided into multiple segments, and then the highest scoring segment is selected based on the semantic similarity of each segment to the query sentence.
    \item \emph{Dense Locators.} To avoid the high computational effort associated with using the candidate window approach, multiple locators are placed after the feature interactions. 
    \item \emph{Single-Shot.} It can be seen as a special case of a candidate window, in which the video is sampled to a fixed size, thus there is only one window without ranking, and the location is predicted by the model only once.
    \item \emph{Reinforcement Learning.} Let the agent learn a strategy that allows it to continuously adjust the location and size of the window in a limited number of steps based on the semantics of the query sentence.
    \item \emph{Boundary Aware.} Boundaries are determined by determining the probability of whether  current time step is the start time or the end time. The motivation is that the start time is the time when the semantic similarity changes from low to high, while the case of  the end time is  opposite.
\end{itemize}

\section{Categories of NLVL}
The existing NLVL can be divided two categories: supervised and weakly-supervised methods. Given there are much more supervised work than weakly-supervised ones, we will pay more attention on the classification of strongly supervised work without further classification of weakly supervised ones.
\subsection{Supervised Methods}
In the NLVL task, the supervision method means to use information about which video the sentence
belongs to, and the specific location of the moment
segment corresponding to the sentence. Therefore, it is relatively simple to converge in training with a higher accuracy.
\subsubsection{Proposal-Based Methods}
\begin{figure} 
\centering 
\includegraphics[width=0.5\textwidth]{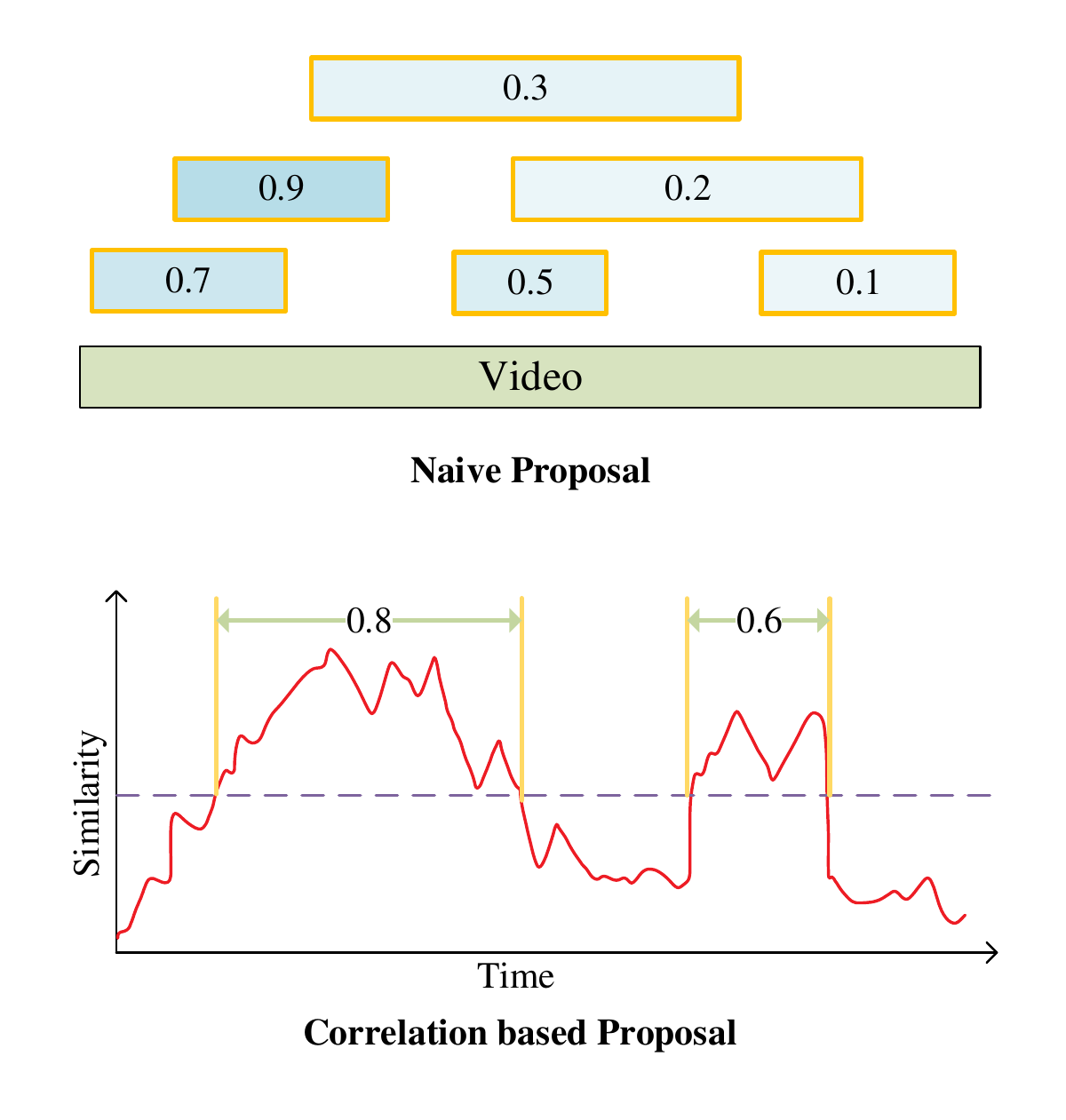}
\caption{Proposal-based methods can be divided into two categories. Naive proposal methods pre-determine the window size and need to handle a large number of candidate proposals. Correlation based methods, on the other hand, reduce the number of candidate proposals.} 
\label{fig:windows} 
\end{figure}
The Proposal-based methods refers to sorting the similarity scores of several different windows of the video and selecting the one with the highest score. It usually consists of the two schemes shown in Figure \ref{fig:windows}. In the naive proposal, windows are first identified and then the similarity scores of each window are calculated. Alternatively, in correlation based proposal, the windows are generated by first calculating the clip-level similarity and then designing the algorithm manually to form a segment. 
Simply dividing the window does not allow for more precise localization of the moment. Therefore many later works fine-tune the boundary on the basis of the window.

Hendricks et al. \cite{anne2017localizing} started the study of NLVL earlier. To address this, they constructed DiDeMo dataset and proposed  Moment Context Network (MCN) model. This model determines the timespan by calculating the semantic similarity between the video segment features of different scales and the sentence's feature. VGG and Glove are used to extract video features and sentence feature, respectively. The local features and global features consider the temporal relationship such as ``The little girl jumps back up after falling" to a certain extent. Later, in order to further consider the 
temporal information in the query sentence, they proposed the TEMPO dataset \cite{hendricks2018localizing} and the Moment Localization with Latent Context (MLLC). The TEMPO dataset emphasizes temporal relationships in events, e.g., a little girl stands up after a fall. Since the window sizes are pre-divided, the loss function is designed to contrast the loss to encourage a greater similarity score for positive windows than for negative windows. It can be simply expressed as
\begin{equation}
    \left [ D(v^+,s)-D(v^-,s)+m \right ] _+,
\end{equation}
where $v^+$ and $v^-$ represents positive and negative video segment features, $s$ represents the query sentence feature, $D$ represents the distance function, and $m$ is the margin. This approach based purely on contrast learning cannot give finer-grained bounds due to the need to pre-segment candidate segments.

Cross-modal Temporal Regression Localizer (CTRL) in Gao et al. \cite{gao2017tall} is also one of the pioneering studies of NLVL. In particular, they use sliding window to generate candidate regions, and design a multi task loss function composed of alignment loss and regression loss. The alignment loss and regression loss optimize the similarity score of the window and the boundary of the window, respectively. This allows to fine tune the window boundary while selecting the window with the highest score. TALL uses C3D as video feature extractor, and uses two different extractors, skip-thought and Glove + LSTM, for sentence features.
For effective inter-modal interaction, they design a module named multi-modal processing, which adds, multiplies, and FC (stitching into fully connection) operations on the feature vectors of two modalities. This module has been used in many subsequent works.
CTRL uses the following losses.
\begin{equation}
            L = L_{align}+\lambda L_{loc}
\end{equation}
\begin{align}
L_{aln}=\frac{1}{N}\sum_{i=0}^{N} [\alpha_{c}\text{log}(1+\text{exp}(-cs_{i,i}))+ \notag\\
 \sum_{j=0,j\neq i}^{N}\alpha_{w}\text{log}(1+\text{exp}(cs_{i,j}))]
\end{align}
\begin{equation}
            L_{r e g} =\frac{1}{N} \sum_{i=0}^{N}\left[R\left(t_{x, i}^{*}-t_{x, i}\right)+R\left(t_{y, i}^{*}-t_{y, i}\right)\right]
\end{equation}
where $ L_{align}$ encourages the model to distinguish between positive and negative video segments with respect to the similarity of the query sentence, and $ L_{r e g}$ fine-tunes the boundaries. $R(t)$ is smooth $L_1$ function.
This loss function that considers both semantic alignment and boundary regression and its variants have been adopted in many subsequent studies.
To mine the matching relationships at a finer granularity, Liu et al. proposed to use the attention mechanism for information mining among modalities. The role of attention in cross-modal interaction is to use information from one modality as a guide to augment information from another modality as a key component. Inspired by this, they proposed two attention schemes, using the sentence feature and the video feature to
guide video features \cite{liu2018attentive} and sentence features \cite{liu2018cross}, respectively.
In addition, in ACRN \cite{liu2018attentive}, they proposed a strategy for modal interaction via outer products. In addition,
Wu et al. \cite{wu2018multi} designed the Multi-modal Circulant Fusion (MCF), which fully interacts the features from two modals on the basis of the circulant matrix. Although this enables element-level interaction of feature vectors from two modalities, it also results in a very large computational increase. 

Previous work has considered sentences containing multiple words as a whole or exploited attention mechanisms to mine the weights of different words. In order to explore the grammatical structure among words in the sentence, Temporal Modular Networks (TMN)  \cite{liu2018temporal} and Temporal Compositional Modular Network (TMCN) \cite{zhang2019exploiting} are proposed, which models the sentence into a tree structure. Ge et al. \cite{ge2019mac} proposed Activity Concepts based Localizer (ACL) to mine the activity concepts in videos and sentences. Specifically, they added the classification layer features of C3D and the verb-obj pair features extracted by Glove as activity concepts to their model. Chen et al. \cite{chen2019semantic} selected the K most common words from the dataset as a lexicon, encoded the sentences as visual semantic concepts by this lexicon with single heat, and then performed similarity matching with video features to find multiple candidate frames that could be of different lengths. Furthermore, they fine-tuned them using regression loss. Jiang  et al.'s Spatial and Language-Temporal Attention (SLTA) \cite{jiang2019cross}, on the other hand, makes use of the object information in the video . Specifically they interact sentence features with object-level visual features extracted by Faster-RCNN \cite{girshick2015fast} in the expectation that fine-grained matching can be achieved.

Considering the naive proposal method provides no flexible window size, it cannot locate moment more accurately. In view of this, Shao et al. \cite{shao2018find} proposed to use the correlation between each clip and sentence to select candidate windows. The  Query-guided Segment Proposal Network (QSPN) proposed by Xu et al. \cite{xu2019multilevel} is also along the same lines, and they also added video captioning as a secondary auxiliary to help training. Another outstanding contribution of QSPN is that they experimentally show earlier interaction for features of two modalities is better than late interaction. 

Ning et al. \cite{ning2021interaction} argued that the traditional windowing approach leads to a certain degree of information loss because it does not cover the query moment exactly. Therefore, they proposed an Interaction-Integrated Network, where the network is able to capture long-range video structure information by overlaying Interaction-Integrated Cells, which is a module that contains a variety of feature interactions including dot product, summation, residual connection, and attention.

\subsubsection{Dense Locators}
\begin{figure} 
\centering 
\includegraphics[width=0.5\textwidth]{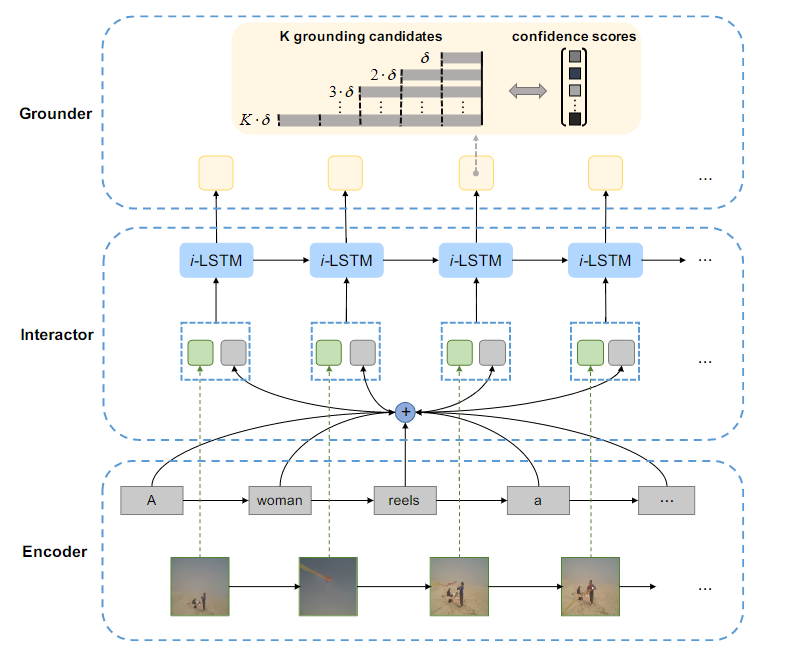}
\caption{Architecture of TGN \cite{chen2018temporally}, note that the scores and bounds of multiple different scale windows are predicted at each time step of the LSTM.} 
\label{fig:tgn} 
\end{figure}
The proposal-based approach requires multiple inputs to process a video, which greatly increases the computational effort. The idea of the dense locators is to move the window to the end, predict multiple boundaries at different moments or different scales, and then rank them based on similarity scores.

To integrate the historical information in the video and complete the input in single pass, Chen et al. designed the  Temporal Ground Net (TGN) \cite{chen2018temporally}, which sequentially scores a set of temporal candidates ended at each frame. This is based on a very bold idea that the video features at each time step after LSTM have temporal information at multiple time scales at the same time. This scheme is counterintuitive, but a large amount of subsequent work has adopted this scheme with good results. For convenience, we call this scheme of predicting multiple windows at one time step the TGN-scheme. Fig.\ref{fig:tgn} illustrates this process. When the video is divided into $T$ moments and $K$ grounding candidates are set, the loss function of training can be expressed as follows, where $c$,$y$ denote the predicted and actual values of similarity, $w$ denote the weight term, respectively.
\begin{align}
    \mathcal{L}_{\mathcal{X}}&=\sum_{(V, S) \in \mathcal{X}} \sum_{t=1}^{T} \mathcal{L}(t, V, S)\\
     \mathcal{L}(t, V, S)&=-\sum_{k=1}^{K} w_{0}^{k} y_{t}^{k} \log c_{t}^{k}+w_{1}^{k}\left(1-y_{t}^{k}\right) \log \left(1-c_{t}^{k}\right)
\end{align}
Cross-Modal Interaction Network (CMIN) \cite{zhang2019cross,lin2020moment} follows the TGN-scheme and applies a syntactic GCN,  a multi-head self-attention,  a multi-stage cross-modal interaction to meet challenges of complex sentence structure, long temporal dependencies, and insufficient cross-modal interactions. 
 In \cite{lin2020moment}, they add a reconstruction loss to strengthen the high-level semantic representation.
The idea of TGN-scheme is also used by DEBUG \cite{lu2019debug}. They divide the existing work into two main categories, the window-based top-down approach and the frame-level
bottom-up approach. As mentioned before, the bottom-up approach is more efficient than the top-down one. Their proposed use of QANet \cite{yu2018qanet} as a backbone network to train the network through three losses of classification, regression, and score (dense head).
On top of DEBUG, Graph-FPN with Dense Predictions (GDP) \cite{chen2020rethinking} further improves the performance by adding graph convolution and feature pyramids to it. 
The idea of dense head is also adopted by Dense Regression Network (DRN) \cite{zeng2020dense}. DRN uses a IoU regression head to explicitly consider the localization quality. 
The TGN-scheme does not stop there. Similar strategy of regressing multiple windows at a single time step is also used by CSMGAN \cite{liu2020jointly}.
CSMGAN constructs both Cross-Modal relation Graph (CMG) and Self-Modal relation Graph (SMG) in the interaction and focus on three different level features, words, paragraphs, and sentences when processing text. By accumulating multiple layers of graphs, CSMGAN is able to fully perform inter- and intra-modal interactions. 

Unlike regressing multiple windows at a single time step, Qu et al. \cite{qu2020fine} truly used sliding windows of different scales over the mixed-modal features, which is more interpretable and have better experimental results. In addition, they used information gate \cite{huang2019attention} to construct Fine-grained Iterative Attention Net-work (FIAN) so that the information from each modal can guide the other modality iteratively. MAN \cite{zhang2019man} and SCDM \cite{yuan2019semantic}  also use rear-mounted multiscale 
locators. Zhang et al. \cite{zhang2019man} used dynamic convolution for modal fusion and designed  Iterative Graph Adjustment Network to enhance the moment representation. In addition, they tried TAN \cite{dai2019tan} for feature extraction and outperformed VGG features on experimental results.
On the basis of MAN \cite{zhang2019man}, Yuan et al. \cite{yuan2019semantic} proposed a novel semantic conditioned dynamic modulation (SCDM) mechanism to model the  temporal convolution procedure based on the sentence semantic.

As opposed to the traditional one-dimensional time window, a two-dimensional temporal feature map was designed by 
Zhang et al. \cite{zhang2020learning} The features at the horizontal and vertical coordinates $i$,$j$ of the temporal map are obtained by averaging pooling from the $i_{th}$ clip to the $j_{th}$ clip of the original video. The most corresponding segment can be found by comparing the similarity between the temporal map and the utterance features. However, it is also a highly computationally intensive scheme, so they use some 
tricks such as sampling to reduce the amount of operations.
Wang et al. \cite{wang2020dual} argued that utilizing both the frame-level
and the candidate-level features will create complementary advantages. Therefore, they proposed Dual Path Interaction Network (DPIN), which utilizes Semantically Conditioned Interaction to accomplish the information transformation between the two levels. 

\subsubsection{Single-Shot}

Many methods claim that they are single-shot, but in fact they move the windows to the end, which greatly reduces the number of calculations. However, they still
require comparative selection among windows. The methods we discuss in this section have only one video input and one output, thus eliminating the need for ranking.

Yuan et al. \cite{yuan2019find} constructed  Attention Based Location Regression (ABLR), which directly incorporates training on attention values in the loss function, and finally uses MLP to generate the final position based on the attention values. The experimental results show that it is better than using features for prediction. Mun et al. \cite{mun2020local} designed a distinct query attention loss function for the semantic features embedded in different parts of a sentence. By drawing on other well-established mechanisms in the field of video understanding, such as position embedding in  \cite{devlin2018bert}, hadamard product  \cite{kim2016hadamard}, and non-local modules \cite{wang2018non}, a fairly high-level accuracy is achieved. 
To make full use of the multi-modal information (visual, motion, audio) of the video to facilitate NLVL, Chen et al. \cite{chen2020learning} proposed the Channel-Gated Modality Interaction model to perform pair-wise modality interaction.
 Chen et al. \cite{chen2020hierarchical} argued that past work has emphasized the importance of action while neglecting the importance of objects in the video. Therefore, they construct a Hierarchical Visual-Textual Graph. HVTG performs visual-textual interaction in both the object and the channel level. In addition, using feature normalization loss \cite{zheng2018ring} makes the model pridect better visual-textual relevance scores. 
\subsubsection{Reinforcement Learning} 
\begin{figure} 
\centering 
\includegraphics[width=0.45\textwidth]{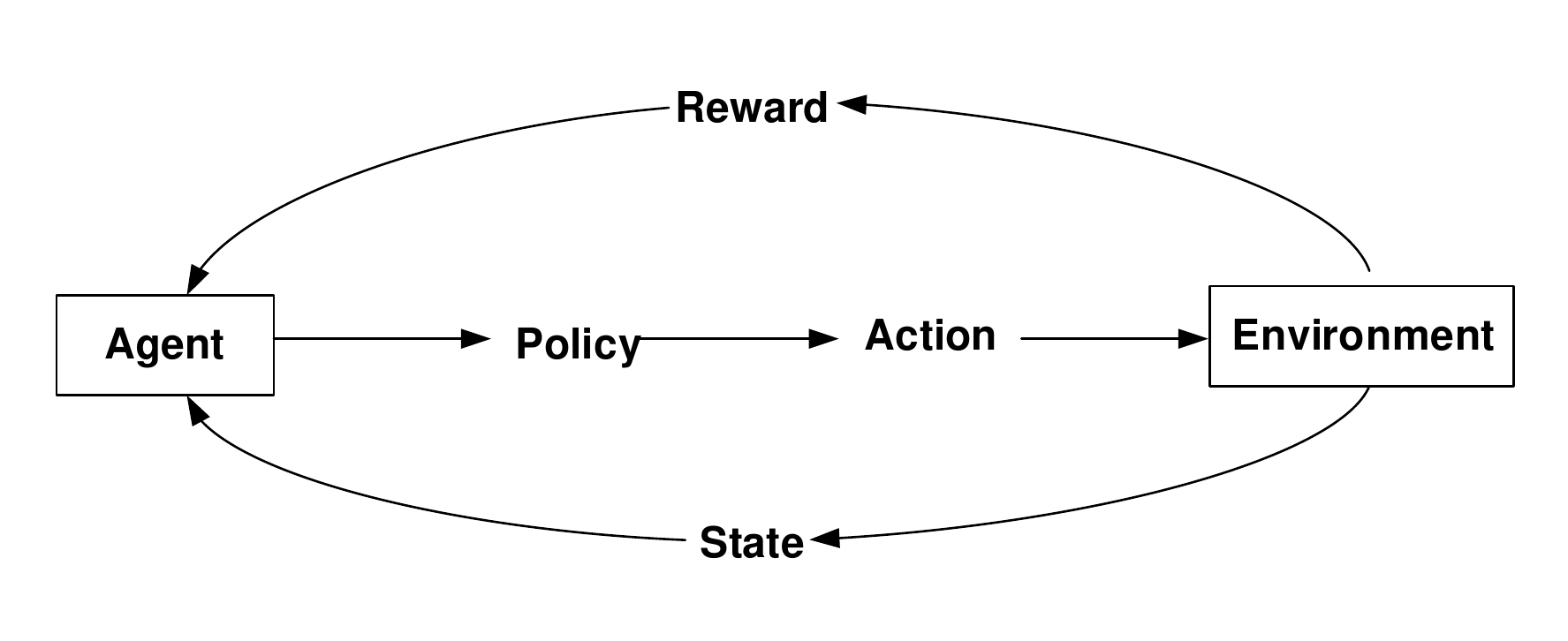}
\caption{Deep reinforcement learning combines the perceptual capabilities of deep learning with the decision-making capabilities of reinforcement learning to continuously adjust the location of boundaries.} 
\label{fig:reinforcement} 
\end{figure}
In recent years, deep reinforcement learning has developed rapidly in the field of video understanding 
\cite{Wang_2018_CVPR2,rao2017attention,han2018reinforcement,torrado2018deep}, and many researchers have focused on its application to NLVL. A typical process of reinforcement learning is shown in Fig.\ref{fig:reinforcement}. An agent obtains the state from the environment and then executes an action according to a certain policy. The action acting on the environment generates a reward, and the agent adjusts the policy according to the reward in order  to maximize the reward. In NLVL task, the agent is often a trainable deep neural network, and the environment refers to video and sentence features. Since the purpose of NLVL is to locate a relevant moment, the action is often  adjusting the observation range. 

Both He et al.  \cite{he2019read} and Wang et al. \cite{wang2019language} used RNN to memorize the state at each time step. Generally speaking, the state here is a vector of features representing high-level semantic information obtained by feeding observations (video features, utterance features, etc.) into a deep neural network. 
These two studies use different strategies of reinforcement learning.
In RWM \cite{he2019read}, they inialized a window and then learned a strategy to move and resize the window to fit the query. 
SM-RL \cite{wang2019language} is different in that it predicts the final bound by selectively observing a sequence fragment. Intuitively, the RWM strategy is easier to understand and obtains better experimental results.
Wu et al. \cite{wu2020tree} proposed the Tree-Structured Policy based Progressive Reinforcement Learning (TSP-PRL), using a two-level policy space to achieve fine-tuning of the boundaries. 
Specifically, the policy branch selects a root policy based on the current state. The root policy is a coarse-grained policy that controls what actions the agent performs, such as scaling, moving left and right, etc. A leaf policy is then selected, which determines the magnitude of the action. The model is able to converge by iteratively optimizing these two tasks.

To align the text and video, Hahn et al. \cite{hahn2019tripping} used gated attention \cite{dhingra2016gated}  showing their model achieves faster localization
than the traditional one.
Cao et al. \cite{cao2020adversarial} proposed an adversarial learning paradigm combining reinforcement learning and moment ranking to solve the NLVL problem. In this way, it is more easier to learn the differences between moments within the same video and can select target segment more efficiently.
Inheriting the advantage of reinforcement learning in avoiding blind selection windows, STRONG (\textbf{S}patio-\textbf{T}emporal \textbf{R}einf\textbf{O}rcement learni\textbf{NG}) \cite{cao2020strong} uses this advantage in selecting important parts from images. In this way, STRONG is able to adjust the regions of interest in both event and spatial dimensions, thus filtering out irrelevant distractions. 
\subsubsection{Boundary Aware}
Chen et al. \cite{chen2019localizing} and Ghosh et al. \cite{ghosh2019excl} used a completely different boundary prediction strategy than before, i.e., a boundary-aware approach. It determines whether a clip is the beginning or the end of an clip based on contextual information.  Thus, finding a segment becomes finding two points, avoiding the candidate window and the corresponding defects. A formal expression of Boundary Aware approach can be shown as in ExCL \cite{ghosh2019excl}:

\begin{align}
\operatorname{span}\left(\hat{t}_{s}, \hat{t}_{e}\right) &=\underset{\hat{t}_{s}, \hat{t}_{e}}{\operatorname{argmax}} P_{\text {start }}\left(\hat{t}_{s}\right) P_{\text {end }}\left(\hat{t}_{e}\right)  \nonumber\\
&=\underset{\hat{t}_{s}, \hat{t}_{e}}{\operatorname{argmax}} S_{\text {start }}\left(\hat{t}_{s}\right)+S_{\text {end }}\left(\hat{t}_{e}\right)  \nonumber \\ 
\text { s.t. } & \hat{t}_{s} \leq \hat{t}_{e}
\end{align}
To predict the starting and ending times, a simple cross-entropy loss function is used for training.
\begin{align}
    L(\theta)=-\frac{1}{N} \sum_{i}^{N} \log \left(P_{\text {start }}\left(t_{s}^{i}\right)\right)+\log \left(P_{\text {end }}\left(t_{e}^{i}\right)\right)
\end{align}

Although both two studies using RNN to mine the temporal information in videos, they differ in the specific processing.
Chen et al. \cite{chen2019localizing} designed self interactor to mine contextual information, which contains a large number ofattention operations, while Ghosh et al. \cite{ghosh2019excl} explored the effect of the structure of three localizers on accuracy. The structure of all three predictors is simple, thus avoiding a large number of calculations.

Wang et al. \cite{wang2020temporally} designed a contextual integration module specifically to capture contextual information to determine whether the current time is the moment boundary.
Rodriguez-Opazo et al. \cite{rodriguez2020proposal} achieved good performance on a simple structure using the designed attention fliter and Kullback-Leibler divergence \cite{hershey2007approximating} loss. To capture the relationships among objects in an image, They later designed a spatial temporal graph algorithm to transfer the information contained in the query sentence to the video feature generation process. To achieve this, their DORi model \cite{rodriguez2021dori} uses both Faster-RCNN and I3D to extract video information.
 Zhang et al. \cite{zhang2020span} added query-guided high-lighting (QGH) strategy on the basis of a QA framework \cite{yu2018qanet} to achieve better results.

\subsection{Weakly-Supervised Methods}
Although there has been a significant amount of work using strongly supervised learning to facilitate NLVL, calibrating the start and end times of moments is labor intensive. On the other hand, due to the diversity and openness of video and textual content, training a robust model to improve our production and life requires a large amount of training data. The motivation for using a weakly supervised approach is to take advantage of the low quality data that is readily available to train the model. Weak supervision in NLVL means that the sentence is only known to which video it belongs, without knowing its exact location.

Duan et al. \cite{duan2018weakly} used iterative training \cite{chidume1987iterative} of the inverse relationship between NLVL and video captions to accomplish the two tasks. In addition, a cross-attention multi-model feature fusion framework is used. In contrast to WS-DEC \cite{duan2018weakly} , Mithun et al. \cite{mithun2019weakly} adopted a simple strategy in  Text-Guided Attention (TGA). That is, the video is pre-segmented into multiple segments of different scales, and the correlation between video segments and the sentence is trained by multiple instance learning. This approach does not easily distinguish segments belonging to the same video, so Gao et al. \cite{gao2019wslln} selected the clip with the largest score as a pseudo-label to reinforce the differences within the video during training.
Since the naive proposal approach brings many drawbacks such as inefficiency and inaccurate alignment, Wu et al. \cite{wu2020reinforcement} tried to apply reinforcement learning \cite{sutton2018reinforcement} to solve the weakly supervised NLVL task. 
Lin et al. \cite{lin2020weakly} pre-selected multiple windows by computing the similarity between the sentence and video clips, and then used the structure of the Encoder-Decoder to reconstruct the masked sentences as a way to train the consistency of sentences and related segment.
Zhang et al. \cite{zhang2020regularized} designed the Regularized Two-Branch Proposal Network based on TAN \cite{zhang2020learning}, considering both intra- and inter-video differences. To enhance the gap between positive and negative samples, a language-aware filter and regularization approach are introduced. Ma et al. \cite{ma2020vlanet} argued that existing weakly supervised methods suffer from dense candidate windows, coarse query representation, and single attention direction, so they designed the Video-Language Alignment Network (VLANet) to select candidate windows and accomplish cross-modal semantic alignment using a bidirectional attention mechanism.
To further increase the contrast between positive and negative samples, Zhang et al. \cite{zhang2020counterfactual} proposed counterfactual contrastive Learning , applying three strategies of feature-level, interaction-level and relation-level to perform counterfactual transformations. In addition, they used gradient \cite{selvaraju2017grad} as one of the considerations when selecting positive and negative samples.
\section{Datasets and Evaluation}
\subsection{Datasets}
There are generally four commonly used datasets, DiDeMo\cite {anne2017localizing},TACoS \cite{regneri2013grounding}, Charades-STA \cite{gao2017tall}, ActivityNet Captions \cite{caba2015activitynet}. However, most methods choose two or three of them for their experiments. Each of these datasets has its own characteristics, which  will be elaborated below.
\subsubsection{DiDeMo}
The Distinct Describable Moments (DiDeMo) \cite{anne2017localizing} dataset has 10,464 videos collected from YFCC100M \cite{thomee2015new}, consisting of 40,543 query pairs. The videos are segmented by machine and human to ensure they are unedited. The videos are between 25 and 30 seconds in length and  cover all aspects of life. To reduce the burden of annotation and to ensure consistent annotation, the videos are divided into multiple 5-second segments, 
so only a rough location of the moment can be provided.
Toward this end, recent supervised studies in rarely use this dataset.
\subsubsection{TACoS}
The TACoS dataset \cite{regneri2013grounding} is re-annotated at multiple levels using crowd-sourcing based on the MPII-Compositive dataset by Rohrbach et al. \cite{rohrbach2012script}
It contains 127 videos and 14,444 query pairs. Gao et al. \cite{Gao_2017_ICCV} divided it into three groups, including a training set, a validation set, and a test set in a 2:1:1 fashion. The annotation provides accurate start and end times, but most methods have relatively low accuracy on this dataset due to the fact that the videos are mainly from the culinary domain and are relatively long (5 minutes on average). 
\subsubsection{Charades-STA}
The Charades-STA dataset is re-labeled by Gao et al. \cite{Gao_2017_ICCV} based on the Charades dataset \cite{sigurdsson2016hollywood}, the original Charades dataset has around 10,000 videos containing actions classified into 157 categories. Gao et al. used a semi-automatic approach to describe them in natural language, so the queries on this dataset are generally simple in syntax. The average length of the videos is around 30 s, while the lengths of the moments are mostly between 6-10 s. In general, this is a relatively stable dataset in terms of distribution.
\subsubsection{ActivityNet Captions}
The ActivityNet Captions dataset is originally developed for video captioning and contains 20,000 untrimmed videos and over 70,000 queries. The video content of the dataset is diverse and open. The average length of the videos is around 2 minutes, while the average length of the moments is around 40s. The length variance of the videos, moments, and query text are all large and therefore difficult.


 \subsection{Evaluation}
The evaluation metric “R@n, IoU=m” is  most commonly used, which is calculated based on the Intersection over Union (IoU). The IoU can be expressed as 
 \begin{equation}
     IoU(P, G)=\frac{|P \cap G|}{|P \cup G|}, 
\end{equation}
 where $P$ and $G$ indicate the predicted timespan and the groundtruth timespan, respectively.
 
$n$ timespans are predicted for each query $N_q$ query sentences. 
 For each query, the symbol $r\left(n, m, q_{i}\right)$) is a indicator, and it is equivalent to one if there is at least one return result when IOU is greater than $m$, otherwise, it is equivalent to zero. Note that some studies consider $r\left(n, m, q_{i}\right)$ to be the proportion of the top $n$ results that reach the IoU threshold $m$.
 
 The metric ``R@n, IoU=m", which is denoteted as$R(m,n)$, is the average of each $r\left(n, m, q_{i}\right)$:
 \begin{equation}
     R(n, m)=\frac{1}{N_{q}} \sum_{i=1}^{N_{q}} r\left(n, m, q_{i}\right).
 \end{equation}



\section{Performance Comparison}
Most of the work chooses 2-3 datasets to validate their ideas, and with the large number of evaluation metrics for NLVL work, we categorized the experimental results by dataset.
Tables [2-4] show the accuracy of each model on the four data sets DiDeMo, TACoS, Charades-STA Caption and ActivityNet. 
The results are obtained from their respective papers. 
We have not ranked these results because these studies differ in their experimental settings. These differences are not only in the hyperparameter settings, such as learning rate, batch size, etc., but also in the features used, which has a significant impact on the experimental results. What's more, additional partitioning of the dataset has been performed in some researches. There are also some papers with highly controversial open source code.

The methods in each table are divided into two main categories according to the supervision. No experiments with weak supervision on the TACoS dataset have been found, so Table\ref{tacos} only has experimental data for the supervised methods. Since there are slight differences in the evaluation metrics used, we used comparable metrics that are employed in most work. Blank cells of the experimental tables indicate experimental results that are not found in the original paper.

Combining the four datasets, the lowest and the highest average metric are
found on dataset TACoS and dataset Charades-STA, respectively. The reason for this difference can be attributed to two main points. On the one hand, it is due to the dataset itself; the queries in dataset TACoS are more detailed, which places a higher demand on the alignment between modalities. On the other hand, much of the work on the Charades-STA dataset uses fine-tuned features  for that dataset, making the representation of the features clearer.

Furthermore, it can be observed that the number of strongly supervised work on the DiDeMo dataset has decreased, while the number of weakly supervised work has gradually increased. This is due to the fact that the dataset is pre-divided into segments, making it easier to  multiple instances learning, which is commonly used in weakly-supervised methods. Also for this reason, the metrics on this dataset will appear with IoU=1.

For the ActivityNet Caption dataset, the relatively large amount of data and the unavailability of a significant portion of the original videos will led to a more stable distribution of experimental metrics as most of the work has been done using the officially provided C3D features. However, this also limits the researchers to  conduct studies on more grassroots work.



\begin{table*}[]
\caption{Performance on DiDeMo}
\label{didemo}
\centering
\begin{tabular}{|l|l|l|l|l|l|l|l|l|l|l|}
\hline
\multicolumn{1}{|c|}{\multirow{2}{*}{model}} & \multicolumn{1}{c|}{\multirow{2}{*}{where}} & \multicolumn{1}{c|}{\multirow{2}{*}{year}} & \multicolumn{4}{c|}{R@1, IoU=} & \multicolumn{4}{c|}{R@5, IoU=} \\ \cline{4-11} 
\multicolumn{1}{|c|}{} & \multicolumn{1}{c|}{} & \multicolumn{1}{c|}{} & 0.5 & 0.7 & 0.9 & 1 & 0.5 & 0.7 & 0.9 & 1 \\ \hline
\multicolumn{11}{|c|}{Supervised Methods}\\\hline 
MCN \cite{anne2017localizing} & ICCV & 2017 &  &  &  & 28.10 &  &  &  & 78.21 \\ \hline
ACRN \cite{liu2018attentive}& SIGIR & 2018 & 27.44 & 16.65 & 16.53 &  & 69.43 & 29.45 & 26.82 &  \\ \hline
TMN \cite{liu2018temporal} & ECCV & 2018 &  &  &  & 22.92 &  &  &  & 76.08 \\ \hline
TGN \cite{chen2018temporally}& EMNLP & 2018 &  &  &  & 28.23 &  &  &  & 79.26 \\ \hline
MLLC \cite{hendricks2018localizing} & EMNLP & 2018 &  &  &  & 28.37 &  &  &  & 78.64 \\ \hline
ROLE \cite{liu2018cross} & MM & 2018 & 29.40 & 15.68 & 15.55 &  & 70.72 & 33.08 & 29.73 &  \\ \hline
TCMN \cite{zhang2019exploiting} & MM & 2019 &  &  &  & 28.90 &  &  &  & 79.00 \\ \hline
SLTA \cite{jiang2019cross} & ICMR & 2019 & 30.92 & 17.16 & 16.43 &  & 70.18 & 33.87 & 30.24 &  \\ \hline
SM-LR \cite{wang2019language} & CVPR & 2019 &  &  &  & 31.06 &  &  &  & 80.45 \\ \hline
MAN \cite{zhang2019man}& CVPR & 2019 &  &  &  & 27.02 &  &  &  & 81.70 \\ \hline 
$\rm I^2N$ \cite{ning2021interaction}& TIP & 2021 &  &  &  & 29.00 &  &  &  & 73.09 \\ \hline 
\multicolumn{11}{|c|}{Weakly-Supervised Methods}\\\hline 
TGA \cite{mithun2019weakly} & CVPR& 2019 & & & &12.19 & & & & 39.74\\ \hline
WSLLN \cite{gao2019wslln} & EMNLP & 2019 & & & & 18.4 & & & &54.4 \\ \hline
RTBPN \cite{zhang2020regularized}& MM & 2020 & & & & 20.79 & & & & 60.26 \\ \hline
VLANet \cite{ma2020vlanet} & ECCV & 2020 & & & & 19.32 & & & &65.68 \\ \hline
\end{tabular}

\end{table*}

\begin{table*}[]
\caption{Performance on TACoS}
\label{tacos}
\centering
\begin{tabular}{|l|l|l|l|l|l|l|l|l|l|l|}
\hline
\multirow{2}{*}{model} & \multirow{2}{*}{where} & \multirow{2}{*}{year} & \multicolumn{4}{c|}{R@1, IoU=} & \multicolumn{4}{c|}{R@5, IoU=} \\ \cline{4-11} 
 &  &  & 0.1 & 0.3 & 0.5 & 0.7 & 0.1 & 0.3 & 0.5 & 0.7 \\ \hline
 \multicolumn{11}{|c|}{Supervised Methods}\\\hline 
CTRL \cite{gao2017tall} & ICCV & 2017 & 24.32 & 18.32 & 13.3 &  & 48.73 & 36.69 & 25.42 &  \\ \hline
ACRN \cite{liu2018attentive} & SIGIR & 2018 & 24.22 & 19.52 & 14.62 &  & 47.42 & 34.97 & 24.88 &  \\ \hline
TGN \cite{chen2018temporally} & EMNLP & 2018 & 41.87 & 21.77 & 18.90 &  & 53.40 & 39.06 & 31.02 &  \\ \hline
MCF \cite{wu2018multi} & IJCAI & 2018 & 25.84 & 18.64 & 12.53 &  & 52.96 & 37.13 & 24.73&  \\ \hline
MAC \cite{ge2019mac} & WACV & 2019 & 31.64 & 24.17 & 20.01 &  & 57.85 & 42.15 & 30.66 &  \\ \hline
ABLR \cite{yuan2019find} & AAAI & 2019 & 34.7 & 19.5 & 9.4 &  &  &  &  &  \\ \hline
SAP \cite{chen2019semantic} & AAAI & 2019 & 31.15 &  & 18.24 &  & 53.51 &  & 28.11 &  \\ \hline
ExCL \cite{ghosh2019excl} & NAACL & 2019 &  & 45.5 & 28.0 & 13.8 &  &  &  &  \\ \hline
SLTA \cite{jiang2019cross} & ICMR & 2019 & 23.13 & 17.07 & 11.92 &  & 46.52 & 32.90 & 20.86 &  \\ \hline
SM-LR \cite{wang2019language} & CVPR & 2019 & 26.51 & 20.25 & 15.95 &  & 50.01 & 38.47 & 27.84 &  \\ \hline
CMIN \cite{zhang2019cross} & SIGIR & 2019 & 32.48 & 24.64 & 18.05 &  & 62.13 & 38.46 & 27.02 &  \\ \hline
SCDM \cite{yuan2019semantic} & CVPR & 2019 &  & 27.64 & 23.27 &  &  & 40.06 & 33.49 &  \\ \hline
DEBUG \cite{lu2019debug} & EMNLP & 2019 & 41.15 & 23.45 & 11.72 &  &  &  &  &  \\ \hline
CMIN \cite{lin2020moment} & TIP & 2020 & 41.73 & 32.35 & 22.54 &  & 69.15 & 50.75 & 32.11 &  \\ \hline
GDP \cite{chen2020rethinking}  & AAAI & 2020 & 39.68 & 24.14 &  &  &  &  &  &  \\ \hline
CBP \cite{wang2020temporally} & AAAI & 2020 &  & 27.31 & 24.79 & 19.10 &  & 43.64 & 37.40 & 25.59 \\ \hline
2D-TAN \cite{zhang2020learning}  & AAAI & 2020 & 47.59 & 37.29 & 25.32 &  & 70.31 & 57.81 & 45.04 &  \\ \hline
PFGA \cite{rodriguez2020proposal} & WACV & 2020 &  & 24.54 & 21.65 & 16.46 &  &  &  &  \\ \hline
DRN \cite{zeng2020dense} & CVPR & 2020 &  &  & 23.17 &  &  &  & 33.36 &  \\ \hline
TripNet \cite{hahn2019tripping}& BMVC & 2020 &  & 23.95 & 19.17 & 9.52 &  &  &  &  \\ \hline
VSLNet \cite{zhang2020span} & ACL & 2020 &  & 29.61 & 24.27 & 20.03 &  &  &  &  \\ \hline
CSMGAN \cite{liu2020jointly} & MM & 2020 & 42.74 & 33.90 & 27.09 &  & 68.97 & 53.98 & 41.22 &  \\ \hline
FIAN \cite{qu2020fine} & MM & 2020 & 39.55 & 33.87 & 28.58 &  & 56.14 & 47.76 & 39.16 &  \\ \hline
DPIN \cite{wang2020dual} & MM & 2020 & 59.04 & 46.74 & 32.92 &  & 75.78 & 62.16 & 50.26 &  \\ \hline
AVMR \cite{cao2020adversarial} & MM & 2020 & 89.77 & 72.16 & 49.13 &  & 94.26 & 83.37 & 64.40 &  \\ \hline
STRONG \cite{cao2020strong}  & MM & 2020 & 90.85 & 72.14 & 49.73 & 18.29 &  &  &  &  \\ \hline
DORi \cite{rodriguez2021dori}  & WACV & 2021 &   & 31.80 & 28.69 & 24.91&  &  &  &  \\ \hline
$\rm I^2N$ \cite{ning2021interaction}  & TIP & 2021 &   & 31.47 & 29.25 & &  & 52.65 & 46.08 &  \\ \hline
\end{tabular}

\end{table*}

\begin{table*}[]
\caption{Performance on Charades-STA}
\label{charades}
\centering
\begin{tabular}{|l|l|l|l|l|l|l|l|l|}
\hline
\multirow{2}{*}{model} & \multirow{2}{*}{where} & \multirow{2}{*}{year} & \multicolumn{3}{c|}{R@1, IoU=} & \multicolumn{3}{c|}{R@5, IoU=} \\ \cline{4-9} 
 &  &  & 0.3 & 0.5 & 0.7 & 0.3 & 0.5 & 0.7 \\ \hline
 \multicolumn{9}{|c|}{Supervised Methods}\\\hline 
CTRL \cite{gao2017tall} & ICCV & 2017 &  & 23.63 & 8.89 &  & 58.92 & 29.52  \\ \hline
ROLE  \cite{liu2018cross}  & MM & 2018 & 25.26 & 12.12 &  & 82.82 & 70.13 & 40.59 \\ \hline
MAC  \cite{ge2019mac} & WACV & 2019 &  & 30.48 & 12.20 &  & 64.84 & 35.13  \\ \hline
MLVI/QSPN \cite{xu2019multilevel}& AAAI & 2019 & 54.7 & 35.6 & 15.8 &  95.6 & 79.4&45.4 \\ \hline
RWM \cite{he2019read} & AAAI & 2019 &  & 36.7 &  &  &  &  \\ \hline
SAP \cite{chen2019semantic} & AAAI & 2019 &  & 27.42 & 13.36 &  &  66.37&38.15 \\ \hline
ExCL \cite{ghosh2019excl}  & NAACL & 2019 & 61.5 & 41.1 & 22.4 &  &  &  \\ \hline
SLTA  \cite{jiang2019cross} & ICMR & 2019 & 38.96 & 22.81 & 8.25 & 94.01 & 72.39 &31.46\\ \hline
SM-LR \cite{wang2019language} & CVPR & 2019 &  & 24.36 & 11.17 &  &   61.25&32.08 \\ \hline
MAN \cite{zhang2019man} & CVPR & 2019 &  & 46.53 & 22.72 &  &   86.23&53.72 \\ \hline
SCDM \cite{yuan2019semantic} & CVPR & 2019 &  & 54.92 & 34.26 &  &   76.50 &60.02\\ \hline
DEBUG \cite{lu2019debug} & EMNLP & 2019 & 54.95 & 37.39 & 17.69 &  &  &  \\ \hline

GDP \cite{chen2020rethinking}  & AAAI & 2020 & 54.54 & 39.47 & 18.49 &  &  &  \\ \hline
CBP \cite{wang2020temporally}  & AAAI & 2020 &  & 36.80 & 18.87 &   & 70.94&50.19 \\ \hline
2D-TAN \cite{zhang2020learning} & AAAI & 2020 &  & 39.81 & 23.25 &  &   79.33&52.15 \\ \hline
TSP-PRL \cite{wu2020tree} & AAAI & 2020 &  & 45.30 & 24.73 &  &  &  \\ \hline
PFGA \cite{rodriguez2020proposal} & WACV & 2020 & 67.53 & 52.02 & 33.74 &  &  &  \\ \hline
LGI \cite{mun2020local} & CVPR & 2020 & 72.96 & 59.46 & 35.48 &  &  &  \\ \hline
DRN \cite{zeng2020dense} & CVPR & 2020 &  & 53.09 & 31.75 &  &   89.06& 60.05\\ \hline
TripNet \cite{hahn2019tripping} & BMVC & 2020 & 54.64 & 38.29 & 16.07 &  &  &  \\ \hline
VSLNet \cite{zhang2020span} & ACL & 2020 & 70.46 & 54.19 & 35.22 &  &  &  \\ \hline
HVTG \cite{chen2020hierarchical}& ECCV & 2020 & 61.37 & 47.27 & 23.30 &  &  &  \\ \hline
PMI \cite{chen2020learning} & ECCV & 2020 & 55.48 & 39.72 & 19.27 &  &  &  \\ \hline
FIAN \cite{qu2020fine} & MM & 2020 &  & 58.55 & 37.72 &  &   87.80 &63.52 \\ \hline
DPIN \cite{wang2020dual} & MM & 2020 &  & 47.98 & 26.96 &  &   85.53&55.00 \\ \hline
AVMR \cite{cao2020adversarial} & MM & 2020 & 77.72 & 54.59 &   & 88.92 & 72.78 &\\ \hline
STRONG \cite{cao2020strong}  & MM & 2020 & 78.10 & 50.14 & 19.30 &  &  &  \\ \hline 
DORi \cite{rodriguez2021dori} & WACV & 2021 & 72.72& 59.65 &40.56 &  &  &   \\ \hline
$\rm I^2N$ \cite{ning2021interaction}  & TIP & 2021 & & 56.61 &34.14 &  &81.48  &  55.19 \\ \hline
\multicolumn{9}{|c|}{Weakly-Supervised Methods}\\\hline 
TGA \cite{mithun2019weakly} & CVPR & 2019 & 32.14 & 19.94 & 8.84 &   86.58 & 65.52 &33.51 \\ \hline
BAR \cite{wu2020reinforcement} & MM & 2020 & 51.64 & 33.98 & 15.97 &  &  &  \\ \hline
SCN \cite{lin2020weakly} & AAAI & 2020 & 42.96 & 23.58 & 9.97 &  95.56 & 71.80&38.87 \\ \hline
RTBPN \cite{zhang2020regularized} & MM & 2020 & 60.04 & 32.36 & 13.24 &  97.48 & 71.85&41.18 \\ \hline
VLANet \cite{ma2020vlanet} & ECCV & 2020 & 45.24 & 31.83 & 14.17 &   95.70 & 82.85 &33.09\\ \hline
CCL \cite{zhang2020counterfactual} & NIPS & 2020 &  & 33.21 & 15.68 &  & 73.50 & 41.87 \\ \hline

\end{tabular}

\end{table*}

\begin{table*}[]
\caption{Performance on ActivityNet Caption}
\label{activity}
\centering
\begin{tabular}{|l|l|l|l|l|l|l|l|l|l|l|}
\hline
\multirow{2}{*}{model} & \multirow{2}{*}{where} & \multirow{2}{*}{year} & \multicolumn{4}{c|}{R@1, IoU=} & \multicolumn{4}{c|}{R@5, IoU=} \\ \cline{4-11} 
 &  &  & 0.1 & 0.3 & 0.5 & 0.7 & 0.1 & 0.3 & 0.5 & 0.7 \\ \hline
 \multicolumn{11}{|c|}{Supervised Methods}\\\hline 
FIFO \cite{shao2018find} & ECCV & 2018 &  & 28.52 & 13.46 &5.21 &  &  &  &  \\ \hline
TGN \cite{chen2018temporally} & EMNLP & 2018 & 70.06 & 45.51 & 28.47 &  & 79.10 & 57.32 & 43.33 &  \\ \hline

MLVI/QSPN \cite{xu2019multilevel} & AAAI & 2019 &  & 45.3 & 27.7 & 13.6 &  & 75.7 & 59.2 & 38.3 \\ \hline
RWM \cite{he2019read}& AAAI & 2019 &  &  & 36.9 &  &  &  &  &  \\ \hline
ABLR  \cite{yuan2019find} & AAAI & 2019 & 73.30 & 55.67 & 36.79 &  &  &  &  &  \\ \hline
ExCL \cite{ghosh2019excl}  & NAACL & 2019 &  & 63.0 & 43.6 & 23.6 &  &  &  &  \\ \hline
CMIN \cite{zhang2019cross} & SIGIR & 2019 &  & 63.61 & 43.40 & 23.88 &  & 80.54 & 67.95 & 50.73 \\ \hline
SCDM \cite{yuan2019semantic} & CVPR & 2019 &  & 55.25 & 36.90 & 20.28 &  & 78.79 &66.84& 42.92 \\ \hline
DEBUG \cite{lu2019debug} & EMNLP & 2019 & 74.26 & 55.91 & 39.72 &  &  &  &  &  \\ \hline
WSLLN  \cite{gao2019wslln} & EMNLP & 2019 & 75.4 & 42.8 & 22.7 &  &  &  &  &  \\ \hline
CMIN \cite{lin2020moment} & TIP & 2020 &  & 64.41 & 44.62 & 24.48 &  & 82.39 & 69.66 & 52.96 \\ \hline
GDP \cite{chen2020rethinking}  & AAAI & 2020 &  & 56.17 & 39.27 &  &  &  &  &  \\ \hline
CBP \cite{wang2020temporally} & AAAI & 2020 &  & 54.30 & 35.76 & 17.80 &  & 77.63 & 65.89 & 46.20 \\ \hline
2D-TAN \cite{zhang2020learning}  & AAAI & 2020 &  & 58.75 & 44.05 & 27.38 &  & 85.65 & 76.65 & 62.26 \\ \hline
TSP-PRL \cite{wu2020tree}  & AAAI & 2020 &  & 56.08 & 38.76 &  &  &  &  &  \\ \hline
PFGA \cite{rodriguez2020proposal} & WACV & 2020 & 75.25 & 51.28 & 33.04 & 19.26 &  &  &  &  \\ \hline
LGI \cite{mun2020local}& CVPR & 2020 &  & 58.52 & 41.51 & 23.07 &  &  &  &  \\ \hline
DRN \cite{zeng2020dense} & CVPR & 2020 &  &  & 45.45 & 24.36 &  &  & 77.97 & 50.30 \\ \hline
TripNet \cite{hahn2019tripping} & BMVC & 2020 &  & 48.42 & 32.19 & 13.93 &  &  &  &  \\ \hline
VSLNet \cite{zhang2020span} & ACL & 2020& 63.16 & 43.22 & 26.16 &  &  &  &  &   \\ \hline
HVTG \cite{chen2020hierarchical} & ECCV & 2020 &  & 57.60 & 40.15 & 18.27 &  &  &  &  \\ \hline
PMI \cite{chen2020learning} & ECCV & 2020 &  & 59.69 & 38.28 & 17.83 &  &  &  &  \\ \hline
CSMGAN \cite{liu2020jointly} & MM & 2020 &  & 68.52 & 49.11 & 29.15 &  & 87.68 & 77.43 & 59.63 \\ \hline
FIAN \cite{qu2020fine} & MM & 2020 &  & 64.10 & 47.90 & 29.81 &  & 87.59 & 77.64 & 59.66 \\ \hline
DPIN \cite{wang2020dual} & MM & 2020 &  & 62.40 & 47.27 & 28.31 &  & 87.52 & 77.45 & 60.03 \\ \hline 
DORi \cite{rodriguez2021dori} & WACV & 2021 &  & 57.89 & 41.49 & 26.41 &  &  &   &   \\ \hline 
\multicolumn{11}{|c|}{Weakly-Supervised Methods}\\\hline
WS-DEC \cite{duan2018weakly}& NIPS & 2018 & 62.71 & 41.98 & 23.34 &  &  &  &  &  \\ \hline
BAR \cite{wu2020reinforcement} & MM & 2020 &  & 53.41 & 33.12 &  &  &  &  &  \\ \hline
SCN \cite{lin2020weakly} & AAAI & 2020 & 71.48 & 47.23 & 29.22 &  & 90.88 & 71.45 & 55.69 &  \\ \hline
RTBPN \cite{zhang2020regularized} & MM & 2020 & 73.73 & 49.77 & 29.63 &  & 93.89 & 79.89 & 60.56 &  \\ \hline
CCL \cite{zhang2020counterfactual}& NIPS & 2020 &  & 50.12 & 31.07 &  &  & 77.36 & 61.29 &  \\ \hline
\end{tabular}

\end{table*}
\section{Conclusion and Perspectives}
In this study, we provide a comprehensive overview of Natural Language Video Localization (NLVL). We first introduce the objective and the basic pipeline of NLVL, and then divide the existing methods into two categories: the supervised
and the weakly supervised methods. Considering there is more work on supervised learning, we further elaborate the classification according to the localization strategy. Finally, we briefly introduce the dataset and evaluation metrics.

NLVL, as an important component of video understanding, has high research and application value. Many researchers have devoted their efforts to moving it step by step from vision to reality. However, NLVL still has a long way to go that 
the metric of accuracy is still difficult to meet the needs of practical applications. In addition, some other metrics should also be considered in further work. For example, what is the time and space performance of the model? Is it robust to the resolution and length of the video? For the former efficiency issue, some work has been considered. For the latter, however, there is a huge gap between theory and practice because the video lengths in currently used datasets are all much shorter than the most common lengths in reality. For example, uniform sampling of videos results in loss of information, and the number of segments containing similar semantics in the same video increases significantly. In addition, current models suffer from a lack of interpretability. How do different localization strategies determine the final boundaries by mixing features or attention scores? How does the proposed novel cross-modal interaction module work? What's more, existing methods use somewhat outdated video features and text features that may limit the expressiveness of the models. Like some recent work on visual or natural language understanding \cite{lu2019vilbert,tan2019lxmert,su2019vl,li2020unicoder,zhou2020unified} has not been given sufficient attention. However, in the near future, we believe that with improvements in AI hardware and software, NLVL will  significantly drive productivity in  many fields, not just high in experimental metrics.


\ifCLASSOPTIONcaptionsoff
  \newpage
\fi

\bibliographystyle{IEEEtran}
\bibliography{IEEEabrv,bib.bib}

\end{document}